%% file: main.tex
\documentclass[10pt,twocolumn,letterpaper]{article}

\usepackage{cvpr}              
\usepackage{adjustbox}
\usepackage{algorithm}
\usepackage{graphics}
\usepackage{multirow}

\usepackage[accsupp]{axessibility} 
\input{preamble}

\definecolor{cvprblue}{rgb}{0.21,0.49,0.74}
\usepackage[pagebackref,breaklinks,colorlinks,citecolor=cvprblue]{hyperref}

\input{macros}

\def\paperID{23} 
\def\confName{CVPR}
\def\confYear{2024}

\title{PP-SAM: Perturbed Prompts for Robust Adaptation of Segment Anything Model for Polyp Segmentation}

\author{Md Mostafijur Rahman\(^1\), Mustafa Munir\(^1\), Debesh Jha\(^2\), Ulas Bagci\(^2\), and Radu Marculescu\(^1\)\\
\(^1\)The University of Texas at Austin\\
\(^2\)Northwestern University\\
{\tt\small \{mostafijur.rahman, mmunir, radum\}@utexas.edu}\\
{\tt\small \{debesh.jha, ulas.bagci\}@northwestern.edu}
}

\begin{document}
\maketitle
\input{sec/0_abstract}    
\input{sec/1_intro}
\input{sec/2_related_works}
\input{sec/3_methodology}
\input{sec/4_experiments}

\input{sec/6_conclusion}
{
    \small
    \bibliographystyle{ieeenat_fullname}
    \bibliography{main}
}

\end{document}

%% file: preamble.tex
\usepackage[dvipsnames]{xcolor}


%% file: macros.tex
\usepackage{times}
\usepackage{multirow}
\usepackage{graphicx}
\usepackage[utf8]{inputenc}
\usepackage{amsmath}
\usepackage{enumitem}
\usepackage{amsfonts}
\usepackage{algorithm}
\usepackage{algpseudocode}
\usepackage{comment}
\usepackage{multirow}
\usepackage{listings}
\usepackage{xcolor}

\definecolor{codegreen}{rgb}{0,0.6,0}
\definecolor{codegray}{rgb}{0.5,0.5,0.5}
\definecolor{codepurple}{rgb}{0.58,0,0.82}
\definecolor{backcolour}{rgb}{0.95,0.95,0.92}

\lstdefinestyle{mystyle}{
    backgroundcolor=\color{backcolour},   
    commentstyle=\color{codegreen},
    keywordstyle=\color{magenta},
    numberstyle=\tiny\color{codegray},
    stringstyle=\color{codepurple},
    basicstyle=\ttfamily\footnotesize,
    breakatwhitespace=false,         
    breaklines=true,                 
    captionpos=b,                    
    keepspaces=true,                 
    numbers=left,                    
    numbersep=5pt,                  
    showspaces=false,                
    showstringspaces=false,
    showtabs=false,                  
    tabsize=2
}

\newcommand{\Tool}{PP-SAM}

%% file: sec/0_abstract.tex
\begin{abstract}
The Segment Anything Model (SAM), originally designed for general-purpose segmentation tasks, has been used recently for polyp segmentation. Nonetheless, fine-tuning SAM with data from new imaging centers or clinics poses significant challenges. This is because this necessitates the creation of an expensive and time-intensive annotated dataset, along with the potential for variability in user prompts during inference. To address these issues, we propose a robust fine-tuning technique, PP-SAM, that allows SAM to adapt to the polyp segmentation task with limited images. To this end, we utilize variable perturbed bounding box prompts (BBP) to enrich the learning context and enhance the model's robustness to BBP perturbations during inference. Rigorous experiments on polyp segmentation benchmarks reveal that our variable BBP perturbation significantly improves model resilience. Notably, on Kvasir, 1-shot fine-tuning boosts the DICE score by 20\% and 37\% with 50 and 100-pixel BBP perturbations during inference, respectively. Moreover, our experiments show that 1-shot, 5-shot, and 10-shot PP-SAM with 50-pixel perturbations during inference outperform a recent state-of-the-art (SOTA) polyp segmentation method by 26\%, 7\%, and 5\% DICE scores, respectively. Our results motivate the broader applicability of our PP-SAM for other medical imaging tasks with limited samples. Our implementation is available at https://github.com/SLDGroup/PP-SAM. 
\end{abstract}

%% file: sec/1_intro.tex
\section{Introduction}
\label{sec:intro}


Deep learning-based algorithms~\cite{ma2023segment,Rahman_2023_WACV,biswas2023polyp,li2023polyp,shaharabany2023autosam,rahman2023gcascade,rahman2024multi,rahman2024emcad,fan2020pranet} have emerged as a promising tool for detecting precancerous lesions during colonoscopy procedures. Recently, a foundational model, namely the Segment Anything Model (SAM), has been introduced for general-purpose semantic segmentation. Several studies explore its zero-shot inference~\cite{ma2023segment,biswas2023polyp,mazurowski2023segment} or fine-tuning~\cite{li2023polyp,shaharabany2023autosam} potential for polyp segmentation. However, when SAM is fine-tuned using data exclusively from one imaging center/clinic, fine-tuning it for different centers/clinics with potentially out-of-distribution data is crucial due to its limited generalizability. Yet, annotating datasets for new centers poses challenges in terms of time, resources, and cost. 

Additionally, the complexity is compounded by the possibility of user prompts being imprecise during inference. Since the prompts used by endoscopists are subjective, there is a chance of variability from the human factors such as fatigue, experience, and number of cases examined in the day. SAM performs poorly when the endoscopists use a larger (inaccurate) bounding box prompt than the region of interest of the polyp. Therefore, it is critical to develop a robust adaptation method that is resilient to inaccurate (perturbed) bounding box prompts. 

To address these issues, we investigate fine-tuning of SAM, namely \textit{PP-SAM}, for polyp segmentation with \textit{variable bounding box prompt perturbations}. By fine-tuning SAM on colonoscopy images, we demonstrate its superior performance for polyp segmentation, showcasing its potential to enhance colorectal cancer screening and diagnosis across diverse clinical settings. Our approach streamlines the time, cost, and resources required for data annotation during fine-tuning, making it effective for multi-center polyp segmentation. Our main contributions are as follows:
\begin{enumerate}
\item \textbf{{PP-SAM framework}}: We introduce \textit{{PP-SAM}}, a new SAM-based robust adaptation framework with limited data for polyp segmentation. We also explore the transfer learning capabilities of different modules of SAM.

\item \textbf{Variable prompt perturbation:} We propose to use a very simple but effective strategy, \textit{variable perturbation of the bounding box prompt}, during fine-tuning to make the model more robust to prompt perturbation.

\item \textbf{Robustness analysis:} We conduct rigorous analyses on the \textit{robustness of zero-shot and few-shot SAM} to bounding box prompt perturbation during inference. Our experimental results show that zero-shot SAM is highly vulnerable to prompt perturbation, hence our proposed PP-SAM with perturbed prompt-based adaptation strategy can significantly improve the robustness of the model during inference.

\end{enumerate}

\vspace{0.1cm}

The remaining of the paper is organized as follows: Section~\ref{sec:methodology} describes our methodology. Section~\ref{sec:related_work} explains relevant prior work. Section~\ref{sec:experiments} presents the experimental setup and results. Finally, Section~\ref{sec:conclusion} concludes the paper. 



\begin{figure*}
  \centering
  \includegraphics[width=0.8\linewidth]{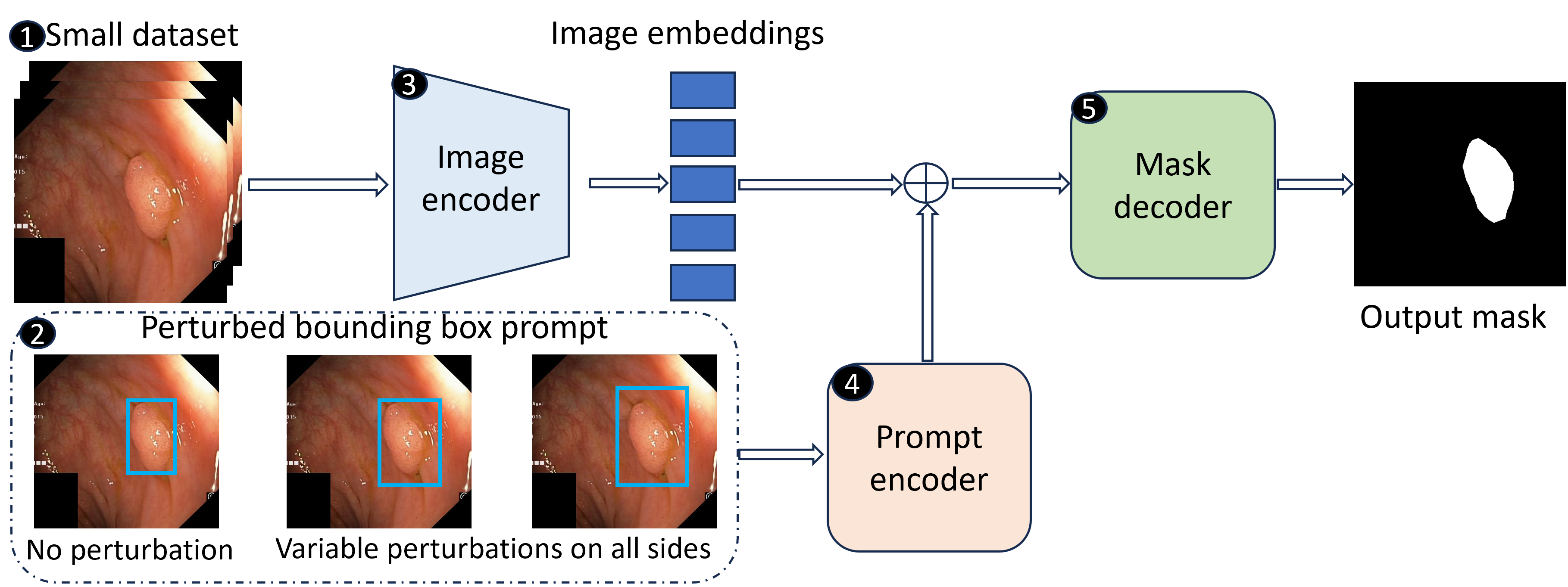}
  \caption{Few-shot fine-tuning pipeline. Here, `no perturbation' represents the bounding box extracted from the original ground truth (GT) masks; `variable perturbations' means extending the bounding box on each side separately.}
  \label{fig:fs_fine_tuining_pipeline}
\end{figure*}

%% file: sec/2_related_works.tex
\section{Related Work}
\label{sec:related_work}

In this section, we describe the related work in three parts: the Segment Anything Model (SAM), SAM in medical image segmentation, and SAM in polyp segmentation.

\subsection{Segment anything model}

In \cite{kirillov2023segment}, the authors of SAM propose a foundation model by introducing a promptable segmentation task, a segmentation model to allow for zero-shot transfer to a variety of tasks and a new dataset for image segmentation. The idea for SAM comes from the natural language processing (NLP) domain, where large language models (LLMs) pre-trained on large datasets have shown strong zero-shot performance \cite{brown2020language}. These types of large models have shown the ability to generalize to tasks and datasets they have not seen during training \cite{bommasani2021opportunities, kirillov2023segment}. The work of SAM \cite{kirillov2023segment} shows that this same type of training on extremely large-scale datasets can be translated to the computer vision domain to segment a variety of different image types, including medical images.

\subsection{SAM in medical image segmentation}

Despite the strong performance of SAM, it fails to perform optimally on out-of-distribution domains, such as medical imaging due to its pre-training on natural images \cite{shaharabany2023autosam, ma2023segment, mazurowski2023segment}. By evaluating SAM on a collection of 19 medical imaging datasets from different anatomies and modalities \cite{mazurowski2023segment}, the authors of \cite{mazurowski2023segment} perform an experimental study to determine the potential of SAM to be applied in medical imaging. The authors find that the performance of SAM based on single prompts is highly dependent on the dataset and the task, thus concluding that SAM shows impressive zero-shot segmentation performance for some medical imaging datasets while performing poorly on others \cite{mazurowski2023segment}.

Auto-SAM \cite{shaharabany2023autosam} replaces SAM's conditioning on either a mask or a set of points with an encoder that operates on the same input image. The introduction of this encoder allows for Auto-SAM \cite{shaharabany2023autosam} to obtain state-of-the-art (SOTA) results on multiple medical image segmentation benchmarks. MedSAM \cite{ma2023segment} goes in another direction by designing a foundation model for medical image segmentation through using a curated dataset with over one million images \cite{ma2023segment}. MedSAM also outperforms existing SOTA foundation models on medical image segmentation and even outperforms some specialized models \cite{ma2023segment}.

\subsection{Polyp segmentation}

Colon polyps are an important precursor to colon cancer, thus the correct segmentation of colon polyps can reduce misdiagnosis of colon cancer \cite{li2023polyp}. Many methods have been proposed for polyp segmentation, but the limited amount of colonoscopy images remains a major challenge. A method like SAM that can perform segmentation without a large amount of data looks thus very appealing. 

In \cite{li2023polyp}, the authors propose Polyp-SAM, which is a fine-tuned version of the SAM model for polyp segmentation. Polyp-SAM achieves SOTA or near SOTA performance on five datasets, thus showing the effectiveness of SAM in medical image segmentation tasks. In \cite{biswas2023polyp}, the authors use a text prompt-aided SAM called Polyp-SAM++, which is shown to be more robust and precise compared to the unprompted SAM \cite{biswas2023polyp, ma2023segment}. The prior work on SAM models (SAM models in medical imaging and SAM models in polyp segmentation), builds upon the original SAM model and uses different methods to achieve better segmentation results than the original SAM for more specialized tasks where pre-training on natural images may not be enough to achieve SOTA results. However, none of these methods considers the real-life inherent inaccuracy (perturbation) in prompts while adapting SAM for polyp segmentation. 

%% file: sec/3_methodology.tex
\section{Methodology}
\label{sec:methodology}

Figure ~\ref{fig:fs_fine_tuining_pipeline} shows our proposed PP-SAM framework. First, we take a small labeled dataset of images as input. Then, we extract the bounding box from the corresponding ground-truth (GT) segmentation mask. Then, we perturb the bounding box using our \textit{variable bounding box prompt perturbation} method. Finally, we use this dataset with GT masks and perturbed bounding box prompts to fine-tune SAM~\cite{kirillov2023segment}. The main components are described next.


\subsection{Prompts}
In this subsection, we describe our proposed variable perturbed prompts for fine-tuning, and perturbation used during inference for robustness analysis.

\subsubsection{Variable perturbed prompts for fine-tuning} 
While SAM can utilize various prompts, we advocate for adapting the bounding box prompt due to its simplicity. Our approach involves fine-tuning SAM for polyp segmentation using a \textit{variable (perturbed) bounding box prompt}. Our variable perturbation extends the bounding box randomly from $0$ to $n$ pixels in all four directions, as in Figure ~\ref{fig:fs_fine_tuining_pipeline} box (2). When we fine-tune SAM with this strategy, the variation in perturbation enhances model robustness against prompt perturbations during real-life inference. 

\subsubsection{Prompts during inference} To assess the robustness of our method, we evaluate the performance of the models with different levels (0, 5, 10, ..., 95, and 100 pixels) of fixed perturbations in the bounding box (on all sides) during inference. For instance, the 10-pixel perturbation during inference means extending the bounding box by 10 pixels on all sides.

\subsection{SAM architecture}
SAM \cite{kirillov2023segment} is a foundational image segmentation model that responds to various prompts (e.g., point, box, mask). While trained on the extensive SA-1B dataset, SAM exhibits robust zero-shot generalization. SAM comprises three key components: image encoder, mask decoder, and prompt encoder. These components are described next.

\subsubsection{Image encoder} 
The SAM image encoder (Figure  \ref{fig:fs_fine_tuining_pipeline} box (3)) is based on a vision transformer (ViT) backbone~\cite{dosovitskiy2020image};
this takes high-resolution (i.e.,
$1024\times1024$) images as input and produces a 16$\times$ downsampled image embedding (i.e., $64\times64$). 

\subsubsection{Prompt encoder} 
The SAM prompt encoder (Figure \ref{fig:fs_fine_tuining_pipeline} box (4)) utilizes two sets of prompts: sparse (bounding boxes, points, text) and dense (masks) prompts. Also, it uses both positional encoding and learned embeddings to encode points and boxes.

\subsubsection{Mask decoder} SAM utilizes a lightweight mask decoder (Figure \ref{fig:fs_fine_tuining_pipeline} box (5)) which consists of a dynamic mask prediction and
an intersection-over-union (IoU) score regression head.



\subsection{Transfer learning} 
We study the transfer learning ability of different components of SAM: image encoder, prompt encoder, and mask decoder. The experimental results in Figure \ref{fig:freezing_exp} reveal that fine-tuning image and prompt encoders suffice; thus, we keep the mask decoder frozen in all our experiments.

\subsection{Limited data settings} 

In this work, we randomly select different small datasets (Figure \ref{fig:fs_fine_tuining_pipeline} box (1)) to fine-tune SAM for polyp segmentation. Specifically, we fine-tune PP-SAM with $k$ randomly selected images, where $k = 1, 5, 10, 20, 50,$ and $100$, as well as the full dataset.

%% file: sec/4_experiments.tex
\vspace{0.1cm}
\section{Experiments}
\label{sec:experiments}
In this section, we describe the datasets, evaluation metrics, implementation details, and experimental results.

\subsection{Datasets}
We use the Kvasir \cite{jha2020kvasir} dataset to fine-tune SAM for few-shot polyp segmentation. This dataset contains 1,000 polyp images. Following~\cite{fan2020pranet}, we adopt the same 900 images from Kvasir as the training set and the remaining 100 images as the testset. To assess the generalizability of our proposed decoder, we use three unseen test datasets, namely ClinicDB ~\cite{bernal2015wm}, EndoScene \cite{vazquez2017benchmark}, and ColonDB \cite{tajbakhsh2015automated}. ClinicDB consists of 612 images extracted from 31 colonoscopy videos. EndoScene and ColonDB consist of 60 and 379 images, respectively. The images of our three unseen testsets significantly differ from trainset as they are collected from separate hospitals/clinics/centers with different acquisition devices and procedures.  

\subsection{Evaluation metrics}
We use DICE similarity scores as evaluation metrics in all our experiments. The DICE similarity score measures the overlap accuracy which is suitable for binary segmentation with imbalanced data. As polyp segmentation is a binary segmentation task with imbalanced polyp (lesion) and background regions, we favor the DICE score for assessing the performance of our PP-SAM on polyp segmentation, where the precise delineation of anatomical structures is crucial for diagnosis and treatment planning. The DICE score $DICE(Y,P)$ of a ground truth mask $Y$ and a predicted mask $P$ is defined in Equation \ref{eq:dice}:
\begin{equation}\label{eq:dice}
DICE(Y, P) = \frac{2 \times \lvert Y \cap P \rvert}{\lvert Y \rvert + \lvert P \rvert}\times100
\end{equation}
\subsection{Implementation details}
We implement and fine-tune our PP-SAM using Pytorch 1.11.0, operating on a single NVIDIA RTX A6000 GPU with 48GB of memory. In our experiments, we set the maximum length for \textit{variable bounding box prompt perturbation}, $n=50$. We resize the images to $1024\times1024$ and re-scale the bounding box to match the new image resolution. We use the AdamW optimizer \cite{loshchilov2017decoupled} with both a learning rate and weight decay rate of 0.0001. We do not use any data augmentations and learning rate schedulers. 

During fine-tuning, we optimize the combined weighted CrossEntropy and mean intersection over union (mIoU) loss function. SAM with ViT-B (SAM-B) is fine-tuned for 100 epochs with a batch size of 1 unless otherwise mentioned; we save the best model based on the DICE score with an inference bounding box perturbation of 30 pixels on all sides. We report the average DICE similarity scores over five runs to evaluate our fine-tuning performance. We calculate the DICE scores using the original resolution of the test images.

\subsection{Results}
To evaluate the performance and robustness of our proposed PP-SAM, we conduct six different sets of experiments as described below.

\begin{figure}
  \centering
  \includegraphics[width=1\linewidth]{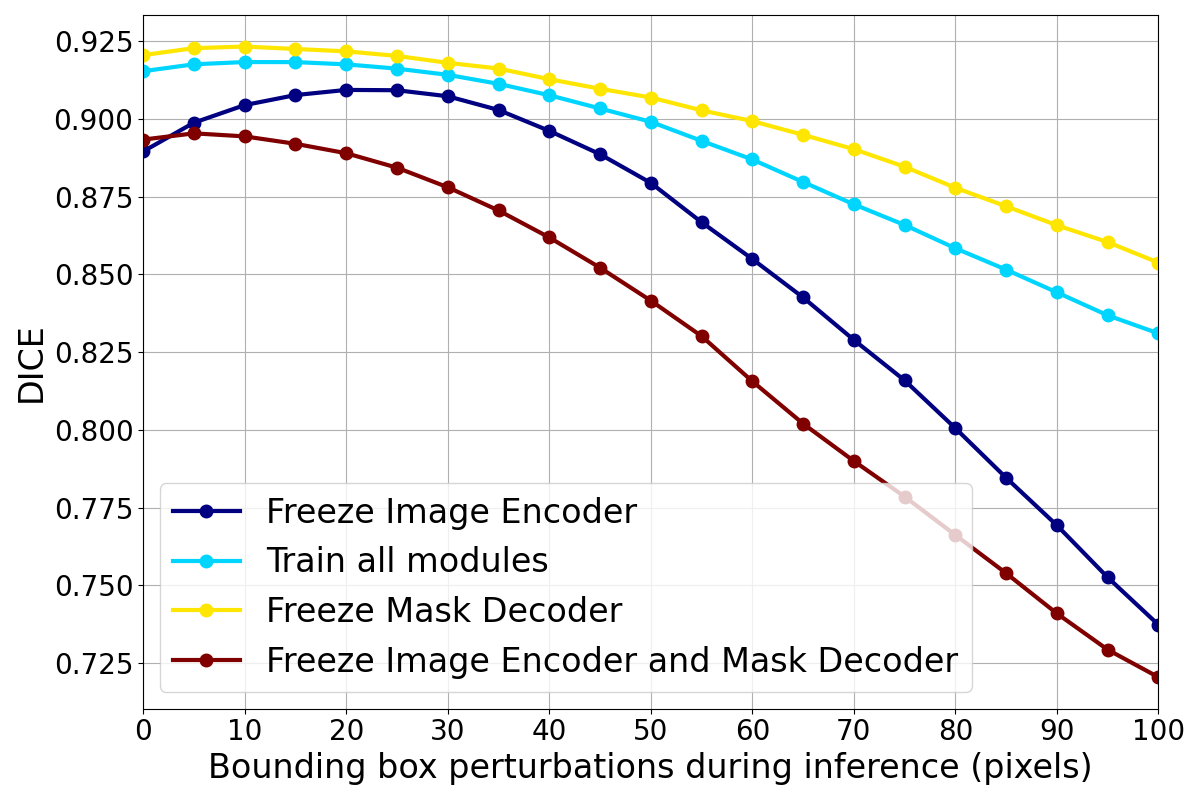}
  \caption{Transfer learning abilities of different modules of SAM on the Kvasir~\cite{jha2020kvasir} testset. As shown, freezing only the mask decoder produces the best results.}
  \label{fig:freezing_exp}
\end{figure}

\begin{figure}
  \centering
  \includegraphics[width=1\linewidth]{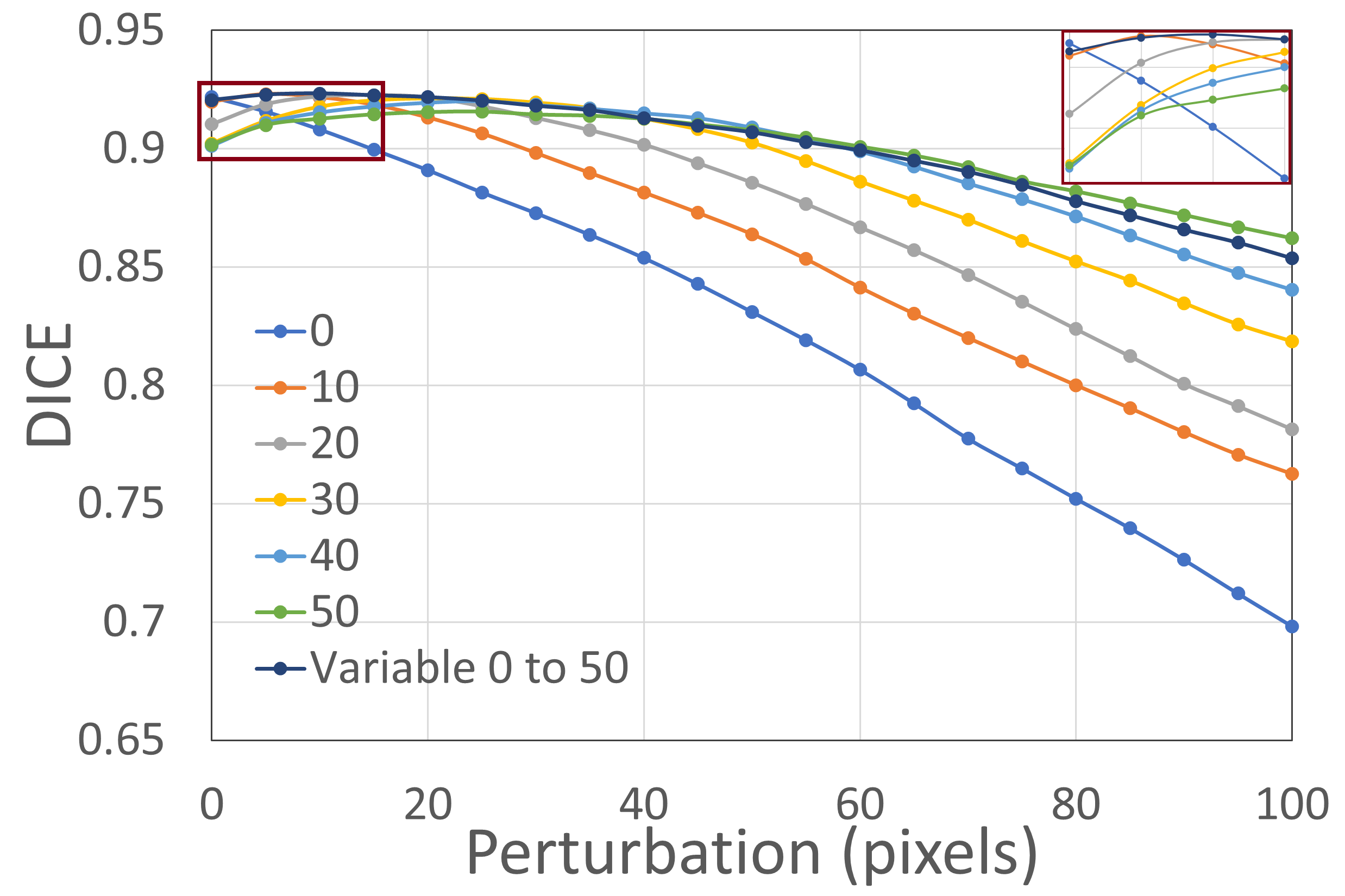}
  \caption{Comparison of different levels of bounding box perturbations during training, on the Kvasir testset. As shown, our variable prompt perturbation produces the overall best results.}
  \label{fig:training_perturbations}
\end{figure}

\subsubsection{Transfer learning capabilities of different modules of SAM} 
We empirically explore SAM's transfer learning capabilities across four distinct experimental configurations. The outcomes of these investigations, as it pertains to the impact of bounding box prompt perturbations during inference, are illustrated in Figure ~\ref{fig:freezing_exp}. In our analysis, the strategic freezing of the mask decoder (i.e., exclusively fine-tuning the image and prompt encoders) emerges as the most effective approach, yielding the highest DICE scores. This superior performance likely stems from the avoidance of overfitting, which can occur when fine-tuning the mask decoder with a limited dataset. Conversely, keeping the image encoder frozen exposes the model to increased vulnerability to prompt perturbation. An even more significant decline in performance is observed when both the image and mask decoders are frozen, underscoring their collective importance in model adaptability. From these insights, we firmly \textit{advocate for freezing only the mask decoder and selectively fine-tuning the image and prompt encoders} to optimize transfer learning efficiency and model robustness.


\subsubsection{Effectiveness of variable bounding box prompt perturbations during fine-tuning} 
In Figure ~\ref{fig:training_perturbations}, we illustrate the impact of different bounding box prompt perturbations during fine-tuning. We assess DICE scores for perturbations of 0 (no perturbation), 10, 20, 30, 40, 50, and random perturbations within 0-50 pixels on all sides of the original bounding box prompt. The results reveal that the models without prompt perturbations during training are susceptible to larger inference perturbations. Resilience to these perturbations improves with larger training perturbations. However, models fine-tuned with variable perturbations (0-50 pixels) demonstrate better overall performance for both small and large inference perturbations. We believe that variable perturbations on different sides during training enhance model robustness against various levels of bounding box prompt perturbations. 


\subsubsection{Learning ability of \Tool{} for polyp segmentation on Kvasir dataset} 
\begin{figure}
  \centering
  \includegraphics[width=1\linewidth]{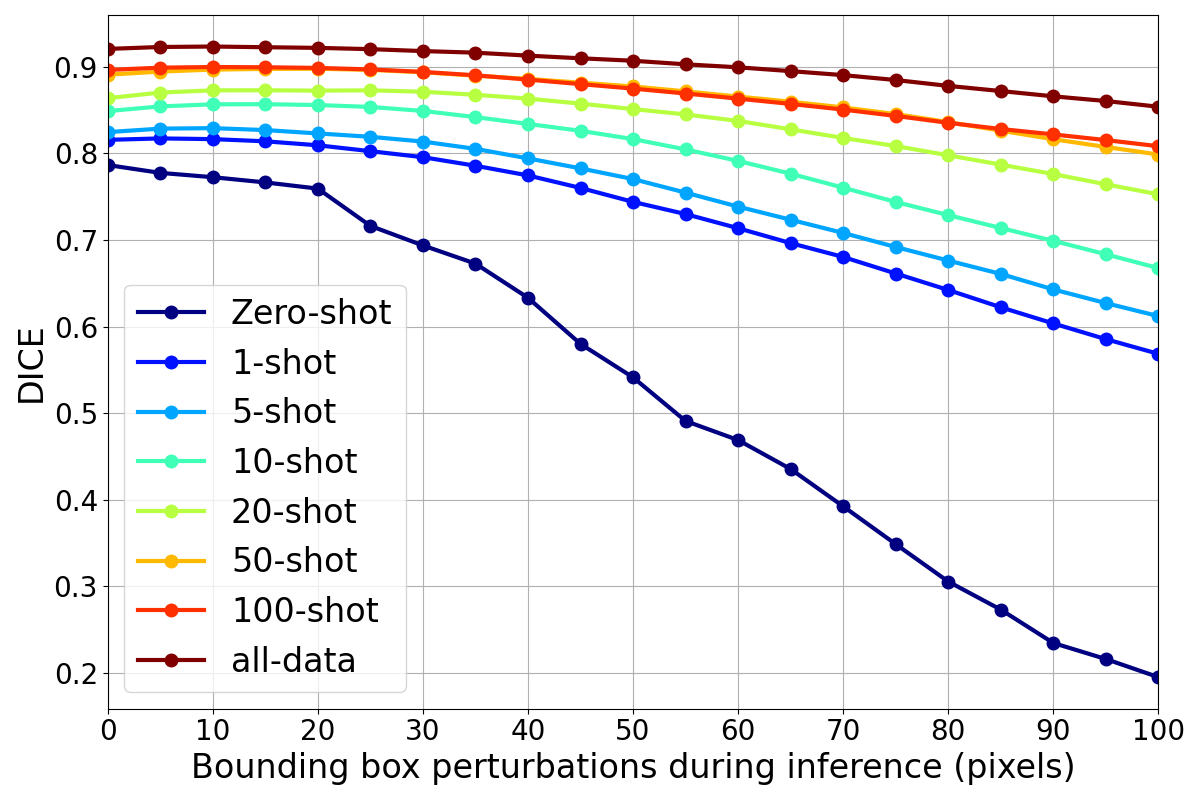}
  \caption{Experimental results on Kvasir testset. All models are trained using the randomly sampled images from the Kvasir trainset. We use our variable perturbed bounding box (in the range of 0 to 50) during training. Also, we keep the mask decoder frozen during these experiments.}
  \label{fig:kvasir_fs_results}
\end{figure}

In Figure ~\ref{fig:kvasir_fs_results}, we present the outcomes of applying both zero-shot and few-shot fine-tuning techniques to the Kvasir dataset during testing. As depicted in this figure, there is a discernible decrease in the DICE scores as the magnitude of bounding box perturbations increases during inference, a trend that aligns with our expectations. Notably, our fine-tuned models demonstrate enhanced durability in the face of these prompt perturbations throughout the inference process. The implementation of our random 1-shot and 50-shot fine-tuning with \textit{variable perturbed bounding box prompts} significantly enhances model robustness, boosting the DICE scores by 37\% and 60\%, respectively, over zero-shot with 100-pixel perturbations on all sides. We can also conclude that the DICE scores improve from 1-shot to 50-shot, with minimal difference beyond 50-shot.

\begin{figure}
  \centering
  \includegraphics[width=1\linewidth]{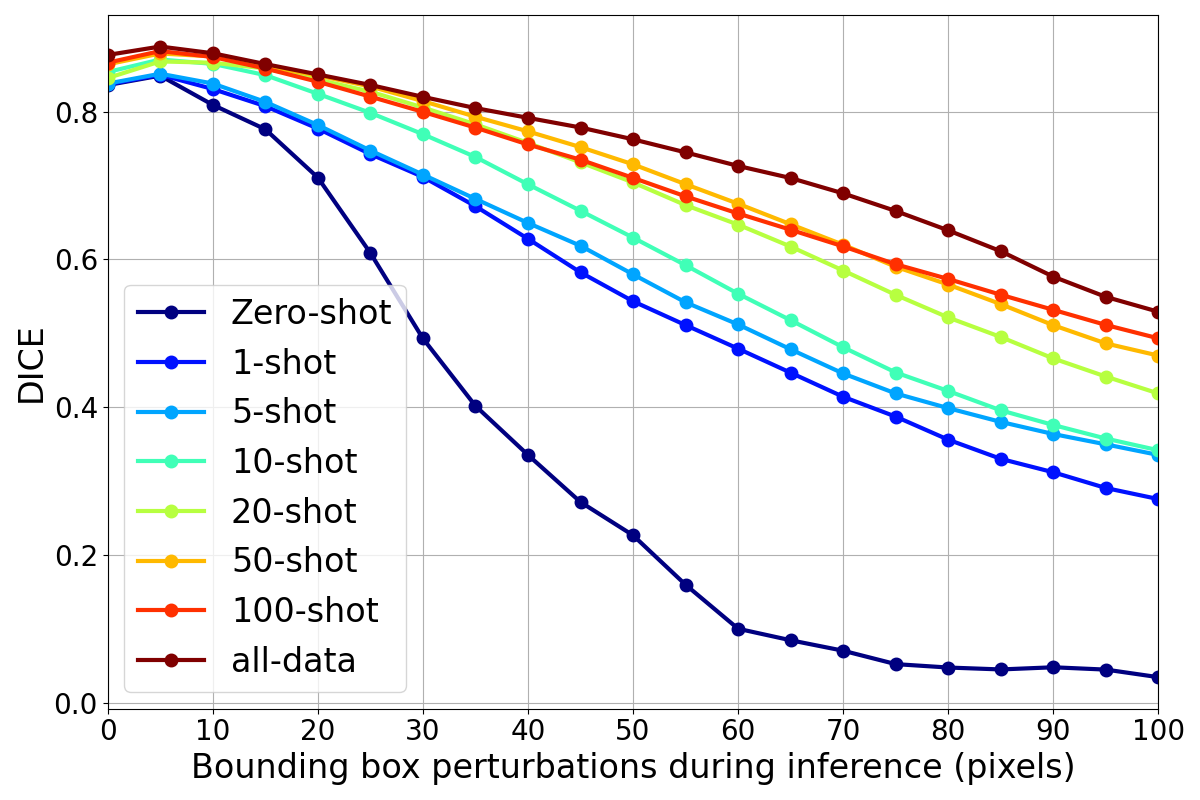}
  \caption{Experimental results of unseen ClinicDB testset. We utilize the models trained using the randomly sampled images from the Kvasir trainset for these experiments.}
  \label{fig:clinicdb_fs_results}
\end{figure}

\begin{figure}
  \centering
  \includegraphics[width=1\linewidth]{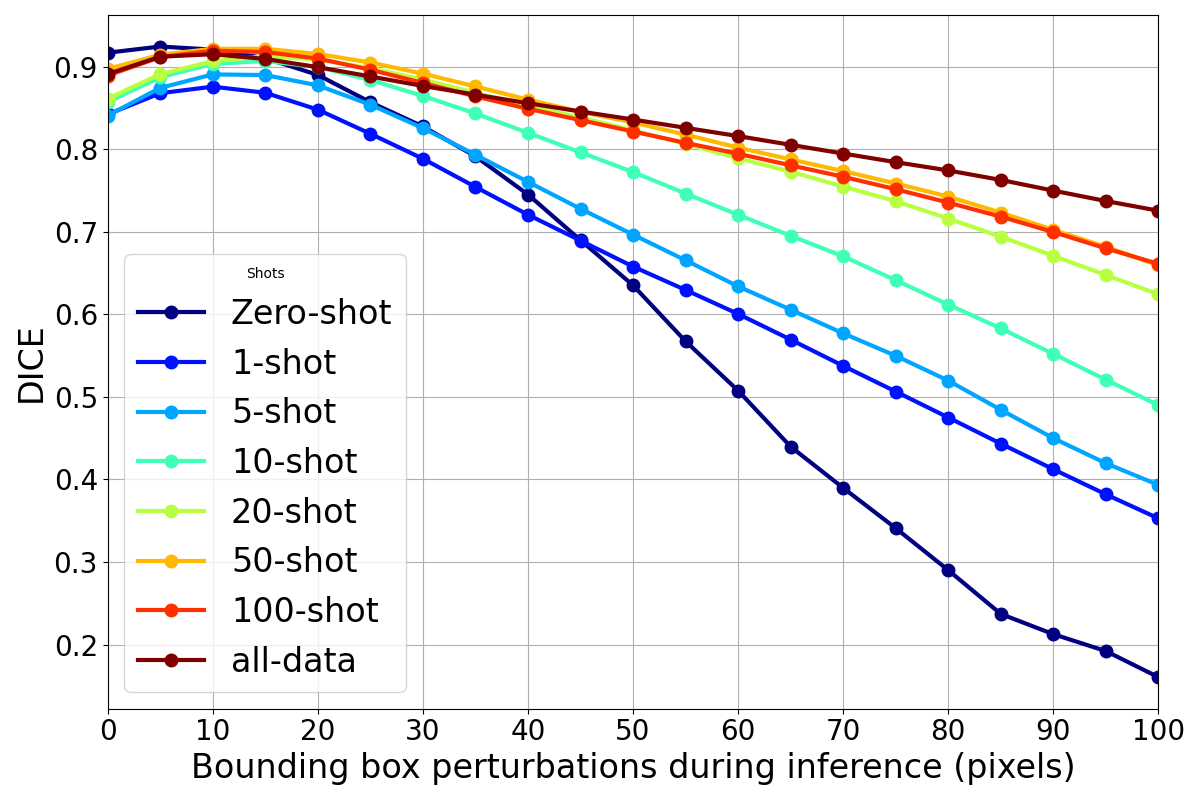}
  \caption{Experimental results on unseen EndoScene testset. We utilize the models trained using the randomly sampled images from the Kvasir trainset for these experiments.}
  \label{fig:cvc300_fs_results}
\end{figure}

\begin{figure}
  \centering
  \includegraphics[width=1\linewidth]{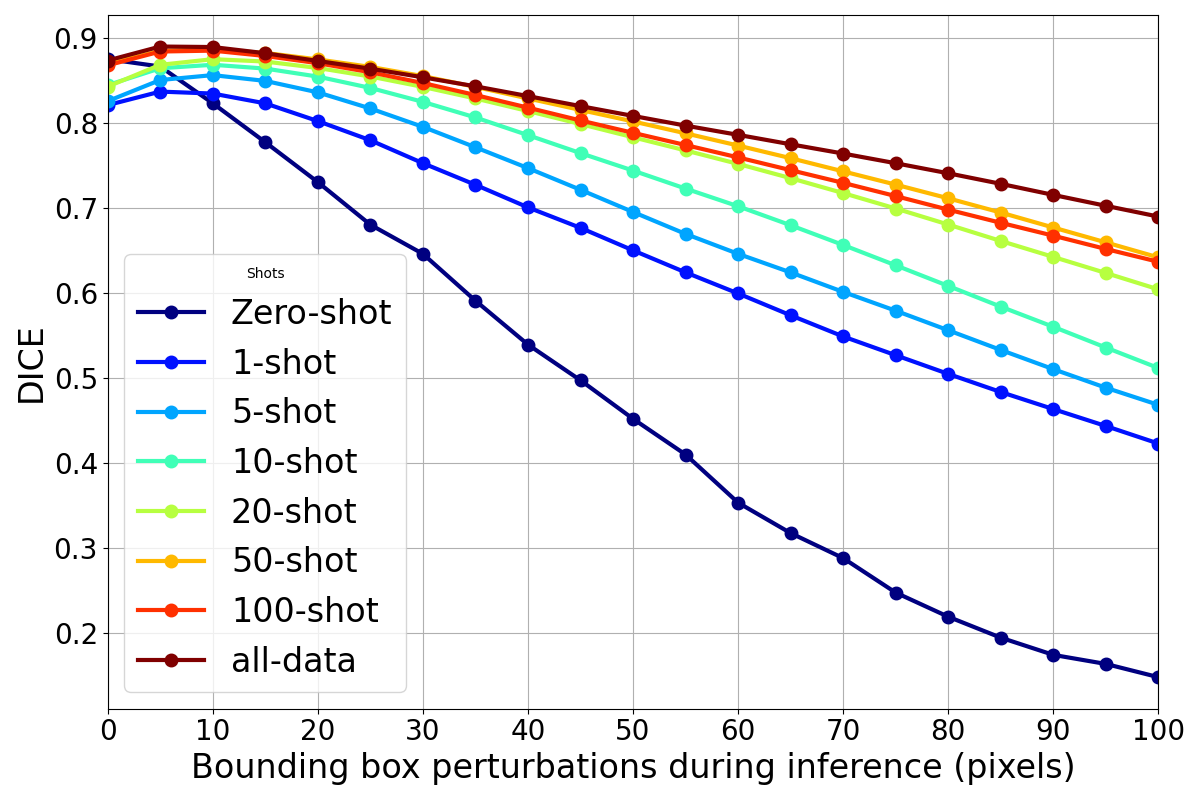}
  \caption{Experimental results on unseen ColonDB testset. We utilize the models trained using the randomly sampled images from the Kvasir trainset for these experiments.}
  \label{fig:colondb_fs_results}
\end{figure}

\subsubsection{Generalizability of \Tool{} for unseen polyp segmentation} Figure ~\ref{fig:clinicdb_fs_results} displays evaluation results of unseen polyp segmentation on ClinicDB testset, where our fine-tuning method shows a significant performance improvement, i.e., 24\% and 43.5\% DICE score increase over zero-shot for 1-shot and 50-shot, respectively. In Figure ~\ref{fig:cvc300_fs_results}, we can see a similar improvement in the DICE scores for the unseen polyp segmentation on EndoScene dataset. More specifically, our 1-shot and 50-shot fine-tuning improves the DICE scores by 19\% and 50\%, respectively, over zero-shot inference with a 100-pixel perturbation. Figure ~\ref{fig:colondb_fs_results} shows the results on the unseen ColonDB dataset, where our 1-shot and 50-shot fine-tuning achieve 27\% and 45\% improvements, respectively, over zero-shot inference with 100-pixel perturbations during inference. We observe minimal improvements in DICE scores after 20 to 50-shot fine-tuning for all the unseen testsets.


\begin{figure}
  \centering
  \includegraphics[width=1\linewidth]{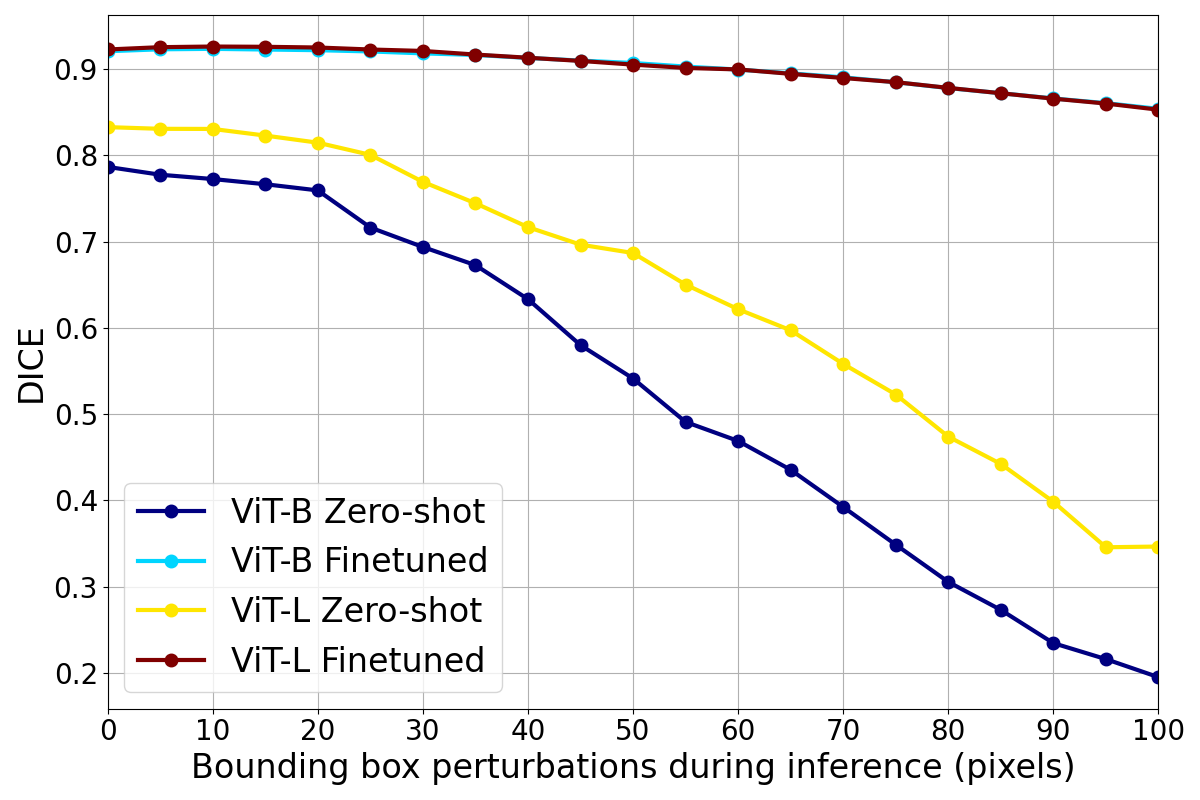}
  \caption{Experimental results of SAM's scalability with ViT encoders on the Kvasir testset. We utilize the models trained using the full Kvasir trainset for these experiments.}
  \label{fig:sam_scalability_results}
\end{figure}

\begin{figure}
  \centering
  \includegraphics[width=1\linewidth]{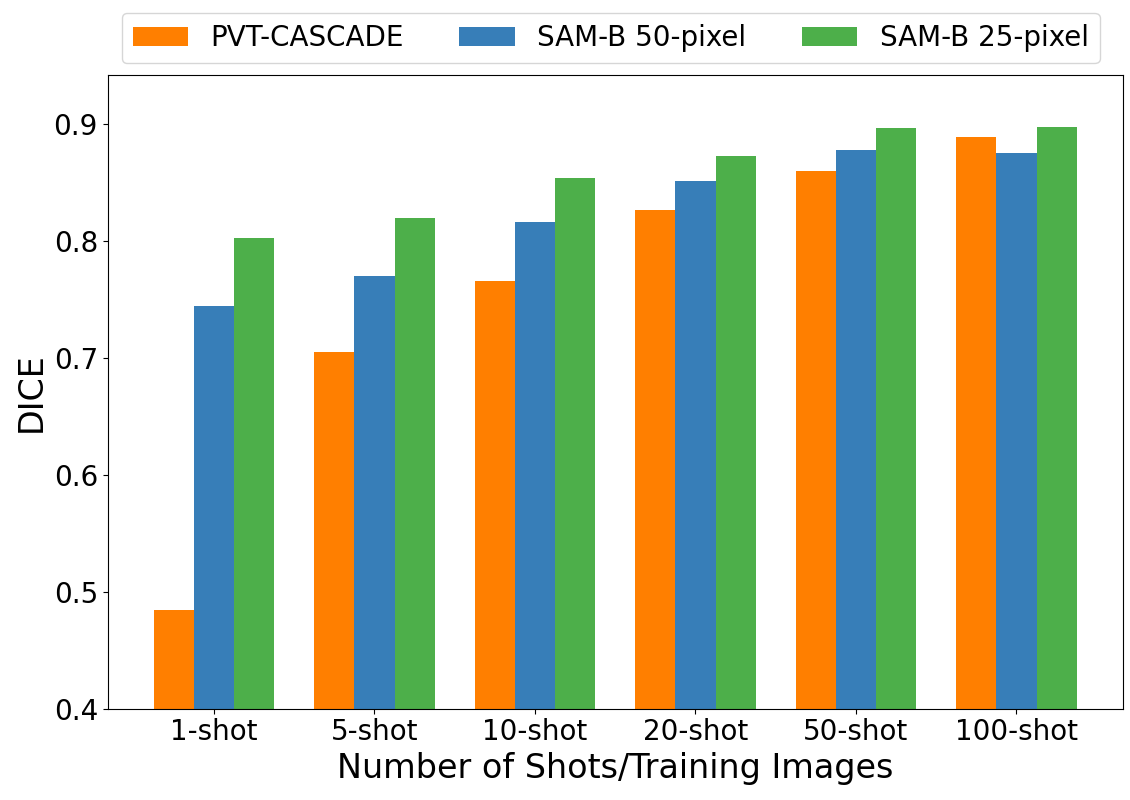}
  \caption{Performance comparison with a SOTA method on the Kvasir testset. We utilize the models trained using randomly selected images from the Kvasir trainset for these experiments. SAM-B 25-pixel and SAM-B 50-pixel are the results of 25 and 50-pixel bounding box perturbations on all sides, respectively, during inference.}
  \label{fig:fs_sota_results}
\end{figure}

\subsubsection{Scalability of SAM with ViT encoders} Figure ~\ref{fig:sam_scalability_results} reports the zero-shot and fine-tuning results of SAM models with ViT-B (Base) and ViT-L (Large) image encoders. From this figure, we can conclude that SAM with ViT-L surpasses ViT-B in zero-shot polyp segmentation. However, the fine-tuned models with both ViT encoders show similar performance. Hence, these results could be used to argue the need for careful consideration when deploying models in real-world scenarios, where perturbations in prompts may occur, and to stress the importance of model robustness to such changes.

\subsubsection{Performance comparison with SOTA} We report the results of our fine-tuned SAM and a SOTA method, PVT-CASCADE \cite{Rahman_2023_WACV} in Figure \ref{fig:fs_sota_results}. We can see from the bar plot that our PP-SAM with variable bounding box perturbations during fine-tuning significantly outperforms PVT-CASCADE for up to 50 shots (i.e., PP-SAM requires less labeled data to achieve closer to optimal performance). More precisely, 1-shot (74.5\%), 5-shot (77\%), and 10-shot (81.6\%) fine-tuning of SAM using our method achieves 26\%, 7\%, and 5\% better DICE scores than PVT-CASCADE with 50-pixel bounding box perturbations during inference. Our method achieves further better performance (i.e., 32\%, 11\%, and 9\% improvement for 1-shot, 5-shot, and 10-shot, respectively) with 25-pixel perturbations. Therefore, our PP-SAM is preferable to non-prompt-based approaches for polyp segmentation involving limited precisely labeled ground truth segmentation masks.

%% file: sec/6_conclusion.tex
\section{Conclusions}
\label{sec:conclusion}

In this paper, we have presented \textit{\Tool{}}, an innovative fine-tuning approach for SAM in polyp segmentation. We introduce a novel concept of \textit{variable perturbations in bounding box prompts} during the training phase, which is aimed at improving the model's robustness against variations and inconsistencies in real-world prompt scenarios. 

The capabilities of \textit{\Tool{}} have been demonstrated to be substantially superior, both in terms of performance enhancements and in maintaining resilience to prompt perturbations, especially when compared to the conventional zero-shot SAM inference methods on publicly available polyp datasets. Our experiments indicate that fine-tuning solely the image and prompt encoder (while freezing the mask decoder) yields superior results. 

While \textit{\Tool{}} currently focuses on binary segmentation and a single bounding box, future work aims to address these limitations. Nonetheless, even in its current state, \textit{\Tool{}} simplifies SAM adoption for new centers/hospitals/clinics, by requiring minimal annotation effort.

\section{Acknowledgements}
This work is supported in part by the NSF grant CNS 2007284, and in part by the iMAGiNE Consortium (https://imagine.utexas.edu/).